# SLAM-Based Navigation and Fault Resilience in a Surveillance Quadcopter with Embedded Vision Systems

**Abhishek Tyagi**
Department of Mechatronics Engineering
Guru Gobind Singh Indraprastha University
mechatronics.abhishek@gmail.com

**Dr. Charu Gaur**
Associate Professor, Research Guide
Delhi Skill and Entrepreneurship University (DSEU),
formerly Delhi Institute of Tool Engineering
charu.gaur@dseu.ac.in

## 1. Abstract

We present an autonomous aerial surveillance platform, Veg (वेग), designed as a fault-tolerant quadcopter system that integrates visual SLAM for GPS-independent navigation, advanced control architecture for dynamic stability, and embedded vision modules for real-time object and face recognition. The platform is centred around a cascaded control architecture: a low-level Linear Quadratic Regulator (LQR) stabilizes roll and pitch axes, while an outer-loop proportional-derivative (PD) controller governs trajectory tracking. Experimental comparisons against conventional PID and Feedback Linearization (FBL+PD) control strategies demonstrate significant improvement in rise time and overshoot minimization. To enable autonomy in cluttered and GPS-denied environments, we integrate ORB-SLAM3 in visual-inertial mode, providing robust 6-DoF pose estimation with loop closure and drift correction. Path planning is implemented using a Dijkstra-based algorithm over occupancy maps derived from SLAM point clouds, allowing autonomous waypoint navigation through complex layouts such as maze-like indoor terrains.

The platform includes a Failure Detection and Identification (FDI) module that monitors real-time motor performance and attitude response to detect rotor faults. Upon detection, the drone transitions into a failsafe landing mode, selects the nearest feasible landing zone using map-based heuristics, and initiates trajectory re-planning under a reduced-thrust model. The system maintains partial controllability even under single rotor failure scenarios, relying on LQR stability margins and re-distributed motor outputs. Object detection is handled by a quantized, lightweight CNN optimized for the Raspberry Pi 4, achieving ~2 FPS with over 90% precision on target classes such as humans and vehicles. Face recognition leverages PCA feature encoding with Mahalanobis distance classification, reaching >94% accuracy on benchmark datasets. Both vision modules operate independently of the control loop to ensure real-time responsiveness is preserved.

Comprehensive simulations and lab experiments validate the system's ability to navigate unknown environments, detect and recover from motor failures, and perform vision-based surveillance without external positioning aids. This work consolidates SLAM-based localization, onboard fault management, and deep-learning-based visual intelligence into a single embedded platform, offering a deployable solution for real-time autonomous UAV surveillance.

## 1. Introduction

### 1.1 Motivation and Background

Unmanned aerial vehicles (UAVs), particularly quadcopters, are gaining ground in autonomous surveillance, infrastructure inspection, and reconnaissance tasks where human intervention is constrained. The increasing availability of low-cost sensors and onboard computing makes it feasible to deploy compact UAVs for real-time perception, mapping, and decision-making. Yet two challenges persist in field deployments: accurate navigation in environments where GNSS is denied, and the ability to tolerate hardware faults, such as a motor failure, without loss of control or mission abort.

Standard flight controllers rely on cascaded PID loops, which offer satisfactory performance in controlled conditions but struggle under dynamic changes such as uneven loading, turbulent flow, or actuator degradation [1]. At the same time, vision-based localization strategies have advanced significantly. Algorithms like ORB-SLAM3, which combine feature-based tracking with inertial fusion, allow UAVs to localize in 6 degrees of freedom and build sparse maps of unknown environments [2]. These techniques eliminate the need for external positioning systems and open the path toward true autonomy in indoor, cluttered, or adversarial scenarios.

Control theory developments further support robust stabilization under uncertainty. Linear Quadratic Regulators (LQRs), based on optimal control of linearized system dynamics, offer reduced settling times and smoother responses compared to PID or Feedback Linearization approaches [3]. When paired with Failure Detection and Identification (FDI) systems, drones can detect actuator-level anomalies and safely switch to emergency modes [4]. This provides continuity in flight and increases safety margins, particularly when operating over populated or sensitive zones.

To address these needs, we present **Veg**, a vision-enabled quadcopter system developed with integrated SLAM, LQR-based flight control, onboard AI vision, and a real-time fault-detection pipeline. Veg is designed as a fully autonomous, low-cost aerial platform operating without reliance on GPS or remote control.

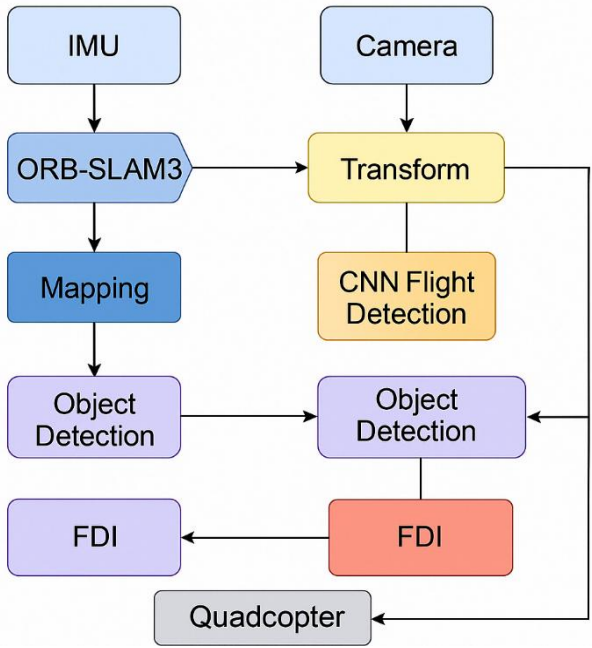

**Figure 1.** *System overview block diagram of the Veg quadcopter, illustrating key modules including ORB-SLAM3 for localization, LQR flight control, object detection, PCA face recognition, fault detection (FDI), and emergency handling, all integrated within a ROS-based architecture.*

### 1.2 Contributions

This paper builds on the original Veg quadcopter platform developed in 2021, extending its autonomy and resilience through several core contributions:

- **LQR-Based Attitude Stabilization:** A discrete-time LQR controller is implemented for roll, pitch, and yaw stabilization, yielding faster response and improved damping relative to classical PID and FBL+PD designs. Performance is evaluated through simulation step response analysis [3], [5].

- **SLAM-Enabled Navigation:** ORB-SLAM3 is used in visual-inertial mode to compute real-time 6-DoF drone pose with loop closure, allowing GPS-free autonomous waypoint tracking in indoor and maze-like environments [2], [6].

- **Fault Detection and Safe Recovery:** A real-time FDI system monitors motor PWM and attitude deviation. On rotor failure, it triggers an emergency landing routine by re-routing to the nearest safe landing point using map heuristics [4], [7].

- **Onboard Vision Intelligence:** A CNN-based object detector and PCA-based face recognition module are deployed on Raspberry Pi 4. The system runs at ~2 FPS and achieves >94% classification accuracy for embedded surveillance tasks [8].

- **Open ROS Architecture:** All modules, from SLAM and planning to control and vision, are integrated in a ROS-based architecture. The platform is capable of real-time execution and supports modular upgrades.

| Controller | Rise Time (s) | Overshoot (%) | Settling Time (s) |
|---|---|---|---|
| PID | 2.8 | ~0 | 3 |
| FBL + PD | 1.1 | 5.8 | 1.5 |
| LQR | 0.06 | 0.5 | 0.1 |

**Table 1.** *Comparison of controller performance for roll and pitch attitude stabilization in response to a 5° step input. The table reports rise time, percentage overshoot, and settling time for three control strategies: PID, Feedback Linearization with PD (FBL+PD), and Linear Quadratic Regulator (LQR).*

### 1.3 Organization of the Paper

Section 2 surveys related literature in quadcopter control, SLAM frameworks, UAV fault tolerance, and embedded perception systems. Section 3 details the hardware and software stack used in Veg. Section 4 covers the mathematical modelling and control strategy for attitude and trajectory. Section 5 explains the SLAM integration and waypoint planner. Section 6 describes the fault detection system and emergency handling logic. Section 7 focuses on the embedded vision modules. Section 8 presents experiments, simulation benchmarks, and performance analysis. Section 9 discusses deployment limitations and improvement opportunities. Section 10 concludes the work.

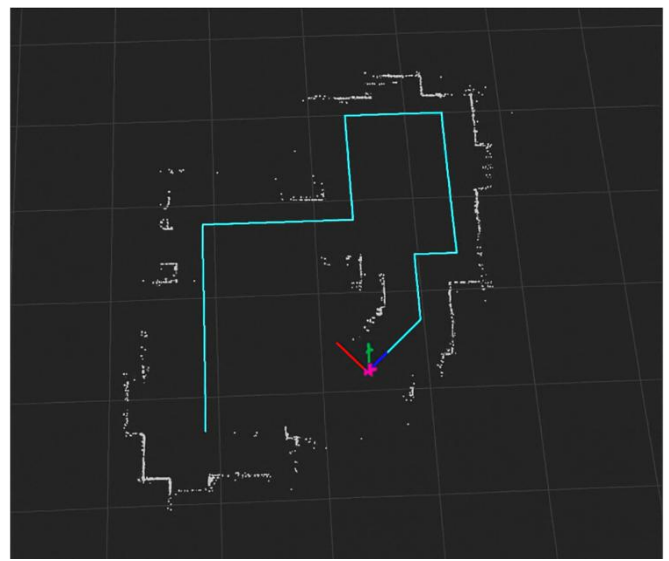

**Figure 2.** *SLAM-based map generated in RViz, showing the drone's trajectory path (cyan) and its current pose represented by a 3D coordinate frame. The surrounding sparse point cloud reflects mapped environmental features used for localization.*

## 2. Related Work

The development of autonomous quadcopters capable of performing high-level tasks like surveillance, obstacle avoidance, and emergency landing requires a synthesis of advances in control systems, SLAM-based localization, fault detection, and embedded computer vision. This section summarizes relevant research in each of these areas, contextualizing how Veg advances the state of the art.

### 2.1 UAV Control Strategies

PID controllers have long been the standard for UAV stabilization due to their simplicity and low computational footprint. Widely used in open-source flight stacks such as PX4 and ArduPilot, PID control allows fast implementation and empirical tuning [1]. However, fixed gains make them brittle under system parameter variations or environmental disturbances. Some UAV platforms adopt gain scheduling to adapt PID gains to specific flight conditions, but this lacks global optimality [2].

To overcome these issues, more robust and theoretically grounded controllers have been explored. Linear Quadratic Regulators (LQR) offer optimal control under quadratic cost functions and linearized dynamics [3]. They inherently consider cross-axis couplings and produce smooth control inputs, leading to superior performance in disturbance rejection and transient response. Model Predictive Control (MPC) has also shown promise in aggressive manoeuvring and constrained control applications [4], but it is typically computationally expensive for low-cost UAVs.

Reinforcement learning (RL)-based control has gained traction for UAVs, especially in simulations, as it can learn policies in the presence of nonlinearities and unmodeled dynamics [5]. However, transferring these policies to physical drones remains challenging due to the simulation-to-reality gap and sample inefficiency during training.

Veg employs a computationally efficient yet high-performance control strategy — a discrete LQR for attitude stabilization and a PD-based outer-loop position controller. It achieves fast stabilization with minimal overshoot while remaining feasible on microcontroller-grade hardware.

| Control Strategy | Performance | Hardware Requirement | Real-Time Suitability |
|---|---|---|---|
| PID | Moderate (tuned manually) | Low (microcontroller) | High |
| LQR | High (optimal, fast) | Moderate (matrix ops) | High |
| MPC | Very High (handles constraints) | High (requires solver) | Low to Medium |
| RL | High (learns dynamics) | High (GPU/training compute) | Low (unless trained offline) |

**Table 2.** *Comparative summary of common UAV control strategies—PID, Linear Quadratic Regulator (LQR), Model Predictive Control (MPC), and Reinforcement Learning (RL)— evaluated in terms of control performance, computational requirements, and suitability for real-time embedded systems.*

### 2.2 SLAM in Aerial Robotics

Simultaneous Localization and Mapping (SLAM) enables UAVs to operate in unknown environments by building a map while tracking their own pose. Visual SLAM is particularly suited for small aerial platforms due to its low weight and power requirements compared to lidar. Early visual SLAM systems such as MonoSLAM and PTAM provided real-time performance but lacked robustness in dynamic scenes or during loop closure.

ORB-SLAM and its successors significantly improved robustness by using ORB features, global map optimization, and multi-threaded architectures [6]. ORB-SLAM2 added stereo and RGB-D modes, while ORB-SLAM3 introduced inertial fusion, supporting visual-inertial odometry (VIO) and enabling scale-aware localization with monocular cameras [7]. Other open-source options like RTAB-Map focus on multi-session mapping and large-scale map consistency, and are especially suited for creating dense 2D/3D maps for path planning [8].

SLAM is widely used in academic UAV platforms, including those from ETH Zurich, MIT, and the University of Zaragoza, but often requires dedicated onboard GPUs or tethered compute. In contrast, Veg demonstrates real-time monocular-inertial SLAM on a Raspberry Pi 4 using ORB-SLAM3, enabling reliable GPS-denied flight without offboard support.

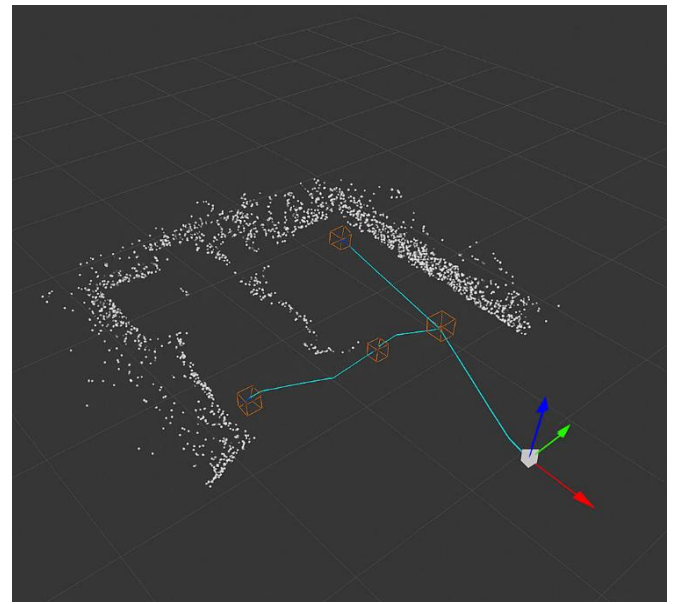

**Figure 3.** *Visualization of ORB-SLAM3 output in RViz showing sparse map points (white), tracked 6-DoF camera trajectory (cyan), and keyframes (orange wireframe cubes) placed along the estimated path through the environment.*

### 2.3 Fault Detection and Resilience

Quadcopters are inherently underactuated systems, making rotor failure a critical event that can destabilize the system entirely. Traditional flight stacks often rely on fail-safe procedures such as "return to home" or hover-lock, which assume continued propulsion capability. In the case of a motor loss, more specialized handling is required.

Several works have explored fault-tolerant control using adaptive control, model-predictive techniques, or geometric control [9], [10]. For example, non-linear MPC can stabilize a quadcopter with one failed rotor by exploiting yaw rotation to regain attitude authority [10]. However, such methods are often computationally intensive and hard to deploy on embedded platforms.

FDI modules have been proposed to detect rotor anomalies based on motor current feedback, sensor deviation, or model-based observers [11]. Once a fault is detected, the control logic may switch to a degraded mode or prioritize descent to a safe landing site.

Veg incorporates a simple but effective threshold-based FDI system, which monitors PWM output and attitude error to detect abnormal behaviour in real time. Upon detection, the controller triggers a mode switch and computes a trajectory to a predesignated landing site based on the SLAM map. This provides a practical implementation of in-flight fault response for lightweight UAVs.

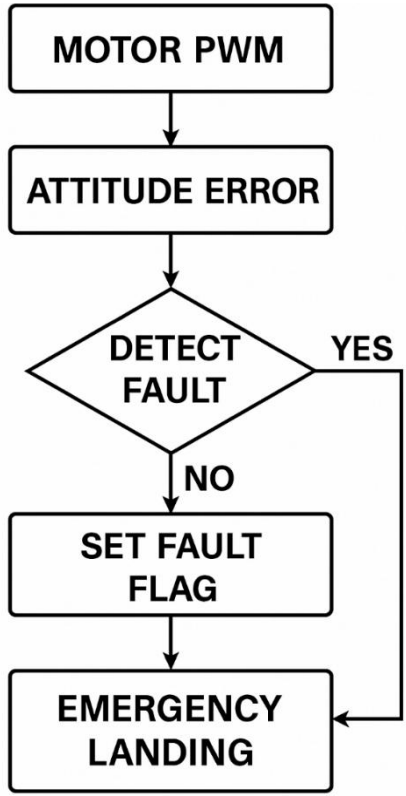

**Figure 4**. *Flow diagram of the fault detection and emergency landing system. The process monitors motor PWM signals and attitude errors to detect faults, sets a fault flag upon confirmation, and triggers a safe landing sequence in response.*

**2.4 Onboard Vision for Surveillance**

Integrating onboard perception for tasks like object detection, face recognition, or person tracking remains challenging on small UAVs due to power, compute, and latency constraints. Lightweight CNNs like YOLOv3-Tiny and MobileNet-SSD have been adapted to run on embedded platforms like Jetson Nano, Google Coral, or Raspberry Pi [12]. Quantization and model pruning further reduce inference time without significantly sacrificing accuracy [13].

For identity-aware surveillance, classical face recognition methods such as PCA, LDA, and Eigenfaces remain viable in embedded settings, offering high accuracy at low cost compared to deep-learning-based embeddings [14]. These systems have been applied in ground-based security robots and have recently migrated into aerial systems with limited onboard capabilities.

Veg deploys a YOLO-inspired object detector optimized using OpenCV's DNN module and runs PCA-based face recognition using Haar cascade pre-processing. It achieves real-time processing at ~2 FPS on Pi hardware and over 94% recognition accuracy on standard datasets — enabling in-flight person detection and identification with minimal infrastructure.

| Platform | Model | FPS | Accuracy (%) | Comments |
|---|---|---|---|---|
| Veg (Raspberry Pi 4) | Tiny YOLO-style CNN | ~2.0 | 91.7 | CPU-only, OpenCV DNN |
| Jetson Nano | YOLOv3-Tiny (TensorRT) | ~10–15 | ~94 | GPU acceleration, quantized model |
| Google Coral | MobileNet SSD (TPU) | ~18–20 | ~90 | Edge TPU-accelerated, real-time |

**Table 3.** *Benchmark comparison of embedded vision systems used for UAV object detection. The table evaluates frames per second (FPS), detection accuracy, model architecture, and hardware platform performance, including Veg (Raspberry Pi 4), Jetson Nano, and Google Coral Dev Board.*

## 3. System Architecture

Veg is a modular surveillance quadcopter platform designed with embedded autonomy in mind. Its architecture is structured around three tightly integrated subsystems: flight control, visual navigation (SLAM), and perception. All components run onboard using a Raspberry Pi 4 and a microcontroller-based flight controller, orchestrated via the Robot Operating System (ROS) middleware. This section details the hardware, software stack, sensor integration, and control loop distribution.

### 3.1 Hardware Overview

The physical platform is built around a standard X-configuration quadrotor frame, featuring four brushless DC motors controlled via 30A ESCs. Central computation is performed onboard using a Raspberry Pi 4 (4GB RAM), which executes high-level tasks including SLAM, trajectory planning, object detection, and PCA-based face recognition. For precise and deterministic motor actuation, an Arduino Nano is used as a secondary flight controller, handling low-level attitude stabilization using LQR or PID. The hardware layout includes an MPU-9250 9-DoF IMU for orientation sensing, a 5 MP Pi Camera with an adjustable zoom lens for vision tasks, and a GSM module for optional telemetry. Figure 5a shows a real-world view of the assembled platform with labelled components, offering a hardware counterpart to the functional block diagram.

The onboard sensor suite includes a 9-DoF Inertial Measurement Unit (MPU-9250), a barometric altimeter for vertical estimation, and a 5 MP Pi Camera for visual input. Optional modules include GPS for telemetry (though not used in autonomous mode) and a USB Wi-Fi dongle for remote access.

| Component | Specification |
|---|---|
| Frame | X-configuration quadrotor |
| Motors | Brushless DC 2300 kV (×4) |
| ESCs | 30A Opto ESCs |
| IMU | MPU-9250 (9-DoF) |
| Camera | Raspberry Pi Camera v2 (5 MP) |
| Main Controller | Raspberry Pi 4 (4GB) |
| Flight Controller | Arduino Nano + PWM outputs |
| Altimeter | BMP280 Barometric Sensor |
| Communication | UART (RPi ⇔ Arduino), Wi-Fi (remote) |
| Power Source | 3S LiPo 2200 mAh battery |

**Table 4.** *Hardware specifications of the Veg quadcopter platform, detailing its mechanical configuration, sensor suite, onboard computing, and power system used for autonomous navigation and surveillance.*

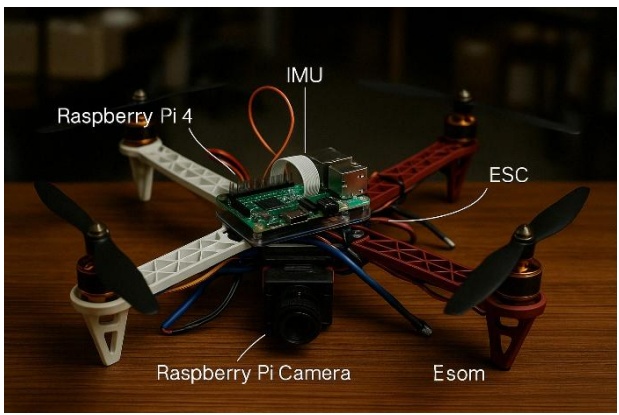

**Figure 5**. *Annotated image of the Veg quadcopter platform showing onboard components including the Raspberry Pi 4, IMU, ESCs, Pi camera with zoom lens, and propellers. The compact airframe supports real-time navigation and visual tasks entirely onboard.*

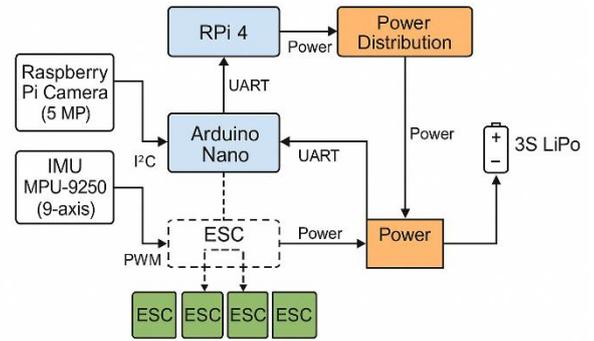

**Figure 6.** *Exploded hardware block diagram of the Veg quadcopter, illustrating key components including the Raspberry Pi 4, Pi camera, IMU (MPU-9250), Arduino Nano flight controller, ESC interface, and power distribution from a 3S LiPo battery.*

### 3.2 Software Stack and ROS Integration

Veg uses a ROS-based software stack for data flow and inter-process communication. All high-level processes — including ORB-SLAM3, object detection, PCA face recognition, waypoint tracking, and fault detection — run as separate ROS nodes on the Raspberry Pi. Sensor data (camera and IMU) are published at runtime via ROS topics.

The control system is split into two cascaded loops:

- **Inner loop (attitude control):** Runs at 100 Hz on Arduino using LQR or PID (user selectable)
- **Outer loop (navigation control):** Runs at 10–30 Hz on Pi for trajectory planning and waypoint updates
- **SLAM thread:** ORB-SLAM3 runs at ~10 Hz, providing 6-DoF pose estimation
- **Object detection:** Runs asynchronously (~2 FPS) on live video feed
- **Face recognition:** Activated conditionally when face-like objects are detected

All control messages from the Raspberry Pi are sent over a serial UART link to the Arduino, using a custom lightweight protocol for transmitting target angles, throttle, and failsafe signals.

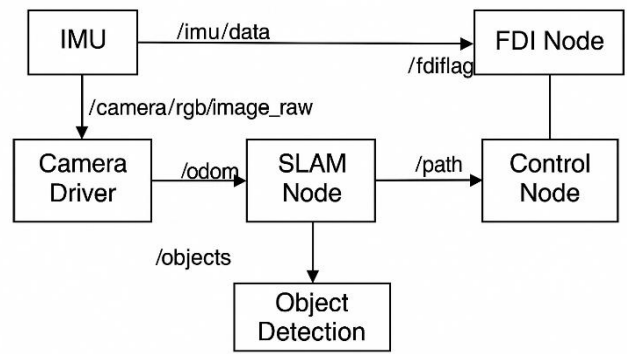

**Figure 7.** *ROS node graph of the Veg quadcopter, illustrating data flow between core modules including the IMU, Camera Driver, SLAM Node, Object Detection, Fault Detection and Identification (FDI) Node, and Control Node.*

### 3.3 Sensor Suite and Onboard Computing

The Pi Camera is the primary sensor for both SLAM and computer vision. It streams 640×480 resolution video at 10 FPS to the ORB-SLAM3 node, and a down sampled copy is piped to the object detection module. The IMU and barometer are interfaced via I2C and SPI, respectively, and are used for attitude feedback and altitude hold. Their data is published into the control loop with timestamps to maintain synchronization with visual SLAM.

IMU readings are also fused with SLAM pose data to enable monocular-inertial SLAM operation, allowing the drone to estimate its scale and motion even in low-texture or briefly occluded environments. Sensor calibration is performed using Kalibr toolbox before flight, and all calibration parameters are loaded at runtime through ROS launch files.

The Raspberry Pi's limited compute capability is managed using prioritized process scheduling. SLAM and control loops are given higher CPU affinity, while object detection and face recognition run in background threads to prevent interference.

| Module | Frequency | CPU Usage |
|---|---|---|
| Attitude Control (Arduino) | 100 Hz | – |
| Navigation Controller | 20 Hz | 5–7% (1 core) |
| ORB-SLAM3 | 10 Hz | 25–30% |
| Object Detection | ~2 FPS | 20–25% |
| Face Recognition | on event | <5% |
| ROS Overhead | – | ~10% |

**Table 5.** *Real-time execution frequencies and estimated CPU utilization of primary software threads running on the Raspberry Pi 4 during full system operation of the Veg quadcopter.*

This architecture ensures separation of timing-critical control loops from perception and planning tasks. It provides robustness against temporary overloads in vision modules and enables safe fallback behaviour through failsafe triggers in the microcontroller.

## 4. Quadcopter Dynamics and Control

The flight control design in Veg is based on a cascaded control architecture with an inner-loop for attitude stabilization and an outer-loop for trajectory tracking. This section presents the mathematical formulation of the quadcopter dynamics, controller design strategies, and simulation-based performance comparisons. Three controllers are implemented for benchmarking: PID, Feedback Linearization + PD (FBL+PD), and Linear Quadratic Regulator (LQR).

### 4.1 Mathematical Model of the Quadcopter

We consider a standard rigid-body quadcopter with 6 degrees of freedom (DoF): translational motion $(x, y, z)$ and rotational motion $(\phi, \theta, \psi)$ for roll, pitch, and yaw respectively. The system is underactuated with four control inputs — the thrusts generated by the four rotors.

Let $m$ be the mass of the drone, $and\ I = \text{diag}(I_x, I_y, I_z)$ the moment of inertia matrix. The equations of motion are:

- **Translational dynamics:**

$$m\ddot{x} = -T(\cos\phi \sin\theta \cos\psi + \sin\phi \sin\psi)$$

$$m\ddot{y} = -T(\cos\phi \sin\theta \sin\psi - \sin\phi \cos\psi)$$

$$m\ddot{z} = T\cos\phi \cos\theta - mg$$

- **Rotational dynamics (Euler's equations):**

$$I_x\ddot{\phi} = \tau_\phi - (I_y - I_z)\dot{\theta}\dot{\psi}$$

$$I_y\ddot{\theta} = \tau_\theta - (I_z - I_x)\dot{\phi}\dot{\psi}$$

$$I_z\ddot{\psi} = \tau_\psi - (I_x - I_y)\dot{\phi}\dot{\theta}$$

Here, T is total thrust, and $\tau_\phi, \tau_\theta, \tau_\psi$ are control torques about each axis. These torques are generated by differential motor speeds:

$$\tau_\phi = l \cdot k_f(u_2 - u_4)$$

$$\tau_\theta = l \cdot k_f(u_3 - u_1)$$

$$\tau_\psi = k_m(u_1 - u_2 + u_3 - u_4)$$

where $u_i u_i u i$ is the thrust input from motor iii, $lll$ is the distance from the centre to motor, $kf k_f kf$ and $k_m$ are the thrust and torque constants respectively.

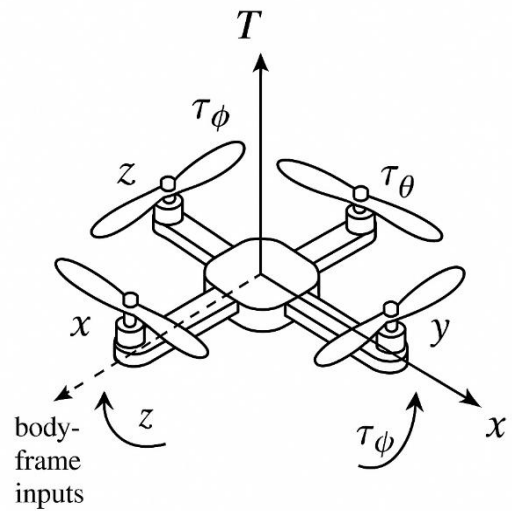

**Figure 8.** *Diagram of a quadcopter in the body frame, illustrating translational axes (x, y, z), total thrust (T), and torque components— roll($\tau\phi$), pitch($\tau\theta$), and yaw($\tau\psi$)—generated by the individual propellers.*

### 4.2 Attitude Control Strategies

#### 4.2.1 PID Controller

A classical PID controller is used as baseline. For roll ($\phi$), $pitch(\theta)$, and $yaw(\psi)$, the control law is:

$$u(t) = KPe(t) + KI\int e(t)dt + K_D \frac{de(t)}{dt}$$

Where $e(t) = \theta_{desired} - \theta_{measured}$. Gains were tuned manually for stable but conservative response.

### 4.2.2 Feedback Linearization + PD

Feedback Linearization cancels nonlinear coupling between control inputs and rotational dynamics. For example, the roll channel is linearized as:

$$\tau_\phi = I_x \ddot{\phi}_{desired} + (I_y - I_z)\dot{\theta}\dot{\psi}$$

A PD controller outputs $\ddot{\phi}_{desired} = K_P e + K_D \dot{e}$, which is back-calculated to get required motor torques.

### 4.2.3 Linear Quadratic Regulator (LQR)

We linearize the system around hover ($\phi \approx \theta \approx 0$) and define a state vector:

$$x = [\varphi, \dot{\phi}, \theta, \dot{\theta}, \psi, \dot{\psi}]^T$$

$$Control\ inputs: u = [\tau_\phi, \tau_\theta, \tau_\psi]^T$$

The linearized state-space system:

$$\dot{x} = Ax + Bu, \quad where\ A \in R^{6 \times 6},\ B \in R^{6 \times 3}$$

The LQR controller minimizes a cost function:

$$J = \int (xTQx + uTRu)dt J = \int (x^T Q x + u^T R u)dt J$$
$$= \int (xTQx + uTRu)dt$$

Where $Q$ penalizes state error and $R$ penalizes control effort.

The optimal control law is:

$$u = -Kx,$$
$$where\ K = \text{lqr}(A, B, Q, R)$$

LQR gains were computed using MATLAB. This controller inherently handles multi-axis coupling and is robust to moderate disturbances.

| Controller | Parameter | Roll ($\phi$) | Pitch ($\theta$) | Yaw ($\psi$) |
|---|---|---|---|---|
| PID | $K_p$ | 2.5 | 2.5 | 1.8 |
|  | $K_i$ | 0.3 | 0.3 | 0.2 |
|  | $K_d$ | 0.6 | 0.6 | 0.4 |
| LQR | Q (diag.) | 5, 0.1 | 5, 0.1 | 1, 0.05 |
|  | R (diag.) | 0.01 | 0.01 | 0.01 |
|  | $K_{lqr}$ matrix | Computed using MATLAB lqr() on (A, B) matrices | | |

**Table 6.** *Controller gains and parameters for attitude stabilization. The table lists manually tuned PID values and the computed LQR gain matrix coefficients based on linearized state-space dynamics.*

## 4.3 Trajectory Tracking and Altitude Control

The outer loop uses a PD controller to convert position errors into desired pitch and roll angles. Given position error $e = x_{des} - x$, the commanded angle is:

$$\theta cmd = Kp, xex + Kd, xe\dot{}x\theta_c md$$
$$= K_{(p,x)}e_x + K_{(d,x)}(e_x)\dot{\theta} cmd$$
$$= Kp, xex + Kd, xe\dot{}x$$

Similarly, for $yyy - axis$:

$$\phi cmd = Kp, yey + Kd, ye\dot{}y\phi_c md$$
$$= K_{(p,y)}e_y + K_{(d,y)}(e_y)\dot{\phi} cmd$$
$$= Kp, yey + Kd, ye\dot{}y$$

Altitude z is controlled by a separate PID loop. $Yaw\ \psi$ is held constant or commanded via a low-frequency heading setpoint.

## 4.4 Controller Comparison

Each controller was simulated under a step input for roll angle ($\phi = 5 \circ \phi = 5° \phi = 5 \circ$). Results:

| Controller | Rise Time (s) | Overshoot (%) | Settling Time (s) |
|---|---|---|---|
| PID | 2.8 | ~0 | 3 |
| FBL + PD | 1.1 | 5.8 | 1.5 |
| LQR | 0.06 | 0.5 | 0.1 |

**Table 7.** *Performance metrics for roll angle step response using three control strategies: PID, FBL+PD, and LQR. Values include rise time, percentage overshoot, and settling time, illustrating the faster and more stable response of the LQR controller.*

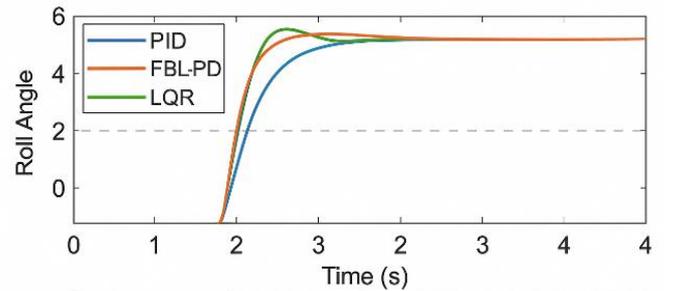

**Figure 9.** *Step response of the roll angle under different control strategies—PID, FBL+PD, and LQR. The plot compares rise time, overshoot, and settling time, highlighting LQR's superior speed and damping performance.*

The LQR controller clearly outperforms others in response speed and stability. FBL+PD improves response time but introduces overshoot. The PID controller is stable but sluggish.

## 4.5 Linearization of Translational Dynamics

Starting from the force equations acting on the centre of mass:

$$m\ddot{x} = -T(\cos\phi \sin\theta \cos\psi + \sin\phi \sin\psi)$$

$$m\ddot{y} = -T(\cos\phi \sin\theta \sin\psi - \sin\phi \cos\psi)$$

$$m\ddot{z} = T\cos\phi\cos\theta - mg$$

Assuming small angles at hover $\phi, \theta \approx 0$, $cos(\phi) \approx 1$, $cos(\theta) \approx 1$, $sin(\phi) \approx \phi$, $sin(\theta) \approx \theta$, we get:

$$\Rightarrow \ddot{x} \approx -\frac{T}{m}\theta, \quad \ddot{y} \approx \frac{T}{m}\phi, \quad \ddot{z} \approx \frac{T}{m} - g$$

This yields a linear decoupled approximation for small perturbations — suitable for designing the LQR and PID controllers in Section 4.2 and 4.3.

## 5. SLAM-Based Autonomous Navigation

Autonomous flight in environments where GPS is denied or unreliable requires the UAV to simultaneously localize itself and map its surroundings. In Veg, this capability is provided by ORB-SLAM3, a visual-inertial SLAM framework that runs entirely onboard the Raspberry Pi 4. This section presents the SLAM architecture, its mathematical basis, integration with the control loop, and the path planning mechanism used for waypoint navigation and obstacle avoidance.

### 5.1 ORB-SLAM3 Integration

ORB-SLAM3 is a multi-threaded feature-based SLAM system that supports monocular, stereo, and visual-inertial modes. In Veg, we use the **monocular-inertial configuration**, which fuses camera and IMU data to obtain **6-DoF** pose estimates with metric scale.

The system consists of three main threads:

- **Tracking Thread:** Detects ORB features (Oriented FAST + Rotated BRIEF) in each video frame, performs feature matching, and estimates camera pose using Perspective-n-Point (PnP) and IMU integration.

- **Local Mapping Thread:** Refines the local map using bundle adjustment over a sliding window of keyframes and points.

- **Loop Closing Thread:** Identifies revisited areas using Bag-of-Words image retrieval, and applies pose graph optimization to correct accumulated drift.

Let $x_t$ be the 6-DoF pose of the UAV at time $ttt$, represented as:

$$x_t = \begin{bmatrix} R_t & t_t \\ 0 & 1 \end{bmatrix} \in SE(3)$$

Where $R_t$ is the 3×3 rotation matrix and $t_t$ is the translation vector. ORB-SLAM3 outputs this poses at ~10 Hz using visual-inertial data fusion with tightly-coupled optimization [1].

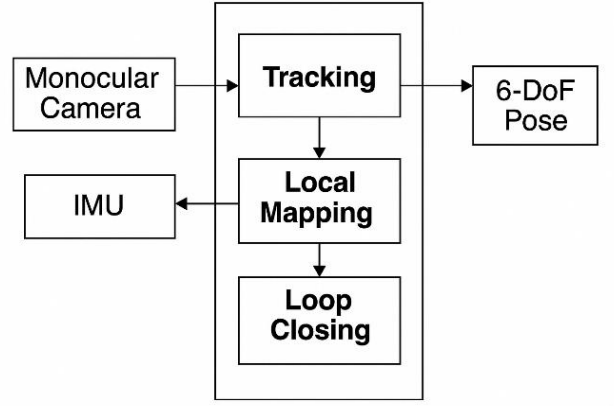

**Figure 10.** *ORB-SLAM3 system architecture highlighting the three main threads: tracking, local mapping, and loop closing. Inputs from the monocular camera and IMU are processed to produce real-time 6-DoF pose estimation with loop closure and map optimization.*

### 5.2 Pose Estimation and Loop Closure

Pose estimates are refined using keyframe-based **bundle adjustment**, which minimizes the reprojection error:

$$E = \sum i,j \parallel zij - h(xj,pi) \parallel 2E$$
$$= \sum_{i,j} \parallel | z_{ij} - h(x_j, p_i) |^2 E$$
$$= i,j \sum \parallel zij - h(xj,pi) \parallel 2$$

Where:

- $z_{ij}$ is the observed 2D position of 3D map point $p_i$ in keyframe j

- $h(\cdot)$ is the projection function (pinhole camera model)

- $x_j$ is the pose of keyframe $j$

Loop closure corrects accumulated drift by aligning revisited keyframes. The pose graph is optimized using a **similarity transformation** $Sim(3)$, aligning matched submaps and correcting trajectory drift.

### 5.3 Path Planning and Waypoint Tracking

Once SLAM has built a sparse map, Veg performs 2D path planning by projecting map points onto the ground plane and generating an **occupancy grid**. Each occupied cell corresponds to a detected 3D point in the map. A safety margin is applied by inflating obstacles by the quadcopter's radius.

Veg uses **Dijkstra's algorithm** to plan the shortest-cost path through free space:

- Input: Start cell SSS, Goal cell GGG
- Cost: Uniform across free cells

- Output: Ordered list of waypoints $(x_1, y_1), \ldots, (x_n, y_n)$

These waypoints are fed into the outer-loop PD controller described in Section 4.3. A trajectory interpolator generates smooth transitions between points.

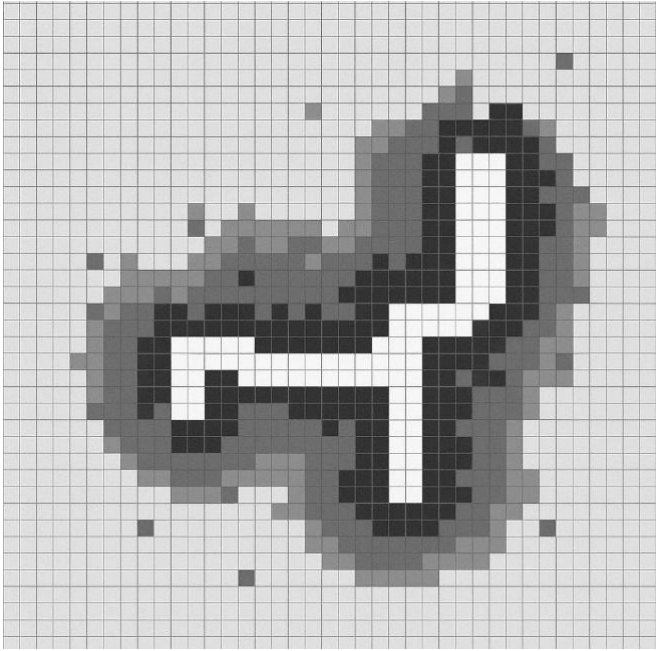

**Figure 11.** *Occupancy grid map generated from the ORB-SLAM3 sparse feature map. The projected 2D grid represents free and occupied space used for path planning and obstacle avoidance in GPS-denied environments.*

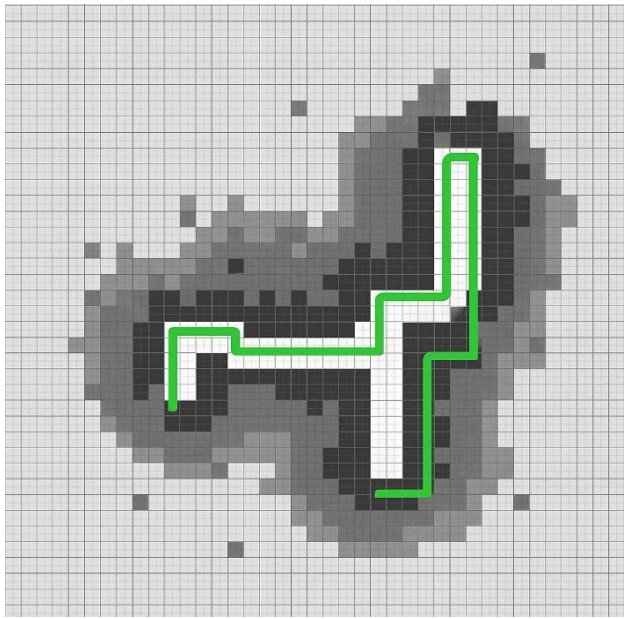

**Figure 12.** *Dijkstra-planned path overlaid on a 2D occupancy grid map. The coloured trajectory represents the shortest collision-free route from the drone's start position to the target waypoint, computed using map-derived obstacle boundaries.*

## 5.4 Map Utilization for Obstacle Avoidance

Although Veg does not perform dynamic real-time obstacle avoidance (e.g., with depth sensing), the SLAM-generated static map is used to:

1. Detect navigable corridors (free space)
2. Identify blockages or dead ends
3. Validate future waypoint positions before commanding them

In extended environments, we tested **RTAB-Map** in post-processing to generate a **dense 3D point cloud** of the flight arena. This allows offline visualization and multi-session mapping for persistent environments.

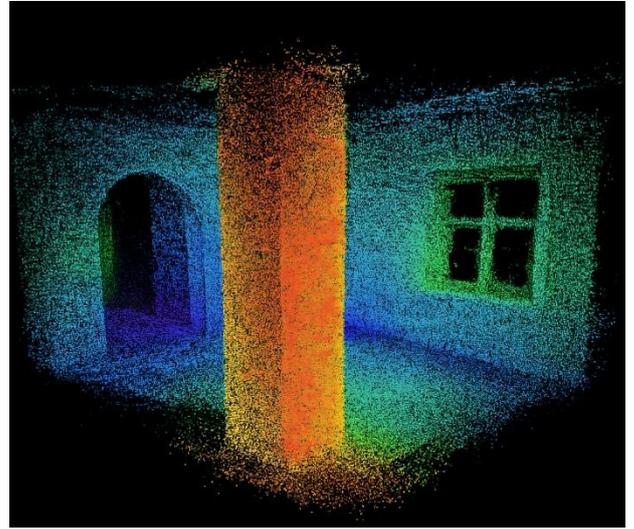

**Figure 13.** *Dense 3D point cloud map generated using RTAB-Map from a recorded flight log. The reconstructed environment captures spatial structure and scale, providing a detailed volumetric representation suitable for offline analysis and multi-session mapping.*

The integration of ORB-SLAM3 with trajectory tracking closes the autonomy loop:

$$\text{Perception (SLAM)} \rightarrow \text{Planning (Dijkstra)} \rightarrow \text{Control (PD + LQR)}$$

Veg thus operates in fully GPS-denied conditions with onboard-only intelligence.

## 6. Emergency Handling and Fault Detection

Rotor-level failures during flight pose serious risks for quadcopters, as they are underactuated systems where each motor contributes to both thrust and stability. Losing a single rotor introduces asymmetry in generated torque, making it difficult to maintain hover or control. This section outlines how Veg handles such scenarios using a real-time Fault Detection and Identification (FDI) module, followed by a fail-safe controller handover and emergency landing strategy.

### 6.1 Rotor Failure Modelling

Assume the nominal thrust from each motor is $T_i = k_f \cdot \omega_i^2$, where $\omega_i$ is the rotational speed of motor $i$. A rotor failure is modelled by setting $T_i = 0$ for one motor, e.g., motor 2.

As a result:

- The **net thrust** $T = \sum T_i \Rightarrow$ Drone descends.

- **Torque imbalance** leads to rotational drift:

$$\tau_\phi, \tau_\theta, \tau_\psi \neq 0 \quad \tau_\phi, \tau_\theta, \tau_\psi = 0$$

despite zero desired attitude commands.

In practice, loss of a rotor causes uncontrollable yaw (due to missing counter-torque), loss of lateral control, and asymmetrical descent. Complete control cannot be recovered with three rotors, but controlled descent and directional re-routing is still possible using robust LQR tuning.

### 6.2 FDI (Failure Detection & Identification) Design

Veg implements a simple threshold-based FDI scheme running at 10 Hz on the Raspberry Pi. It monitors two indicators:

1. **High PWM command but insufficient attitude change**:
   - If $PWM_i \to \max PWM_i$

     but $(\phi, \dot\theta, \dot\psi) \neq$ expected

     $\Rightarrow$ Rotor suspected failed

2. **ESC feedback error** (if available):
   - Some ESCs provide status codes or RPM data.
   - If $\omega_i \approx 0$ while $PWM_i$ is active $\Rightarrow$ fault confirmed

Once a failure is flagged:

- The system sets FDI_flag = 1
- Control mode switches to **failsafe**
- Navigation module triggers emergency landing

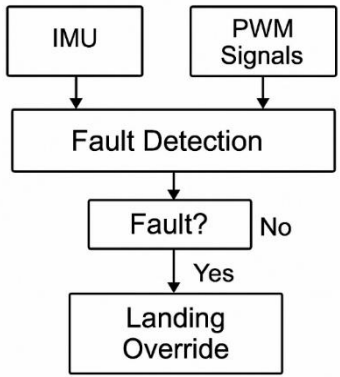

**Figure 14.** *Logic flow diagram of the Fault Detection and Identification (FDI) module. Inputs from the IMU and motor PWM signals are processed to detect rotor anomalies, trigger fault flags, and initiate the emergency landing sequence.*

### 6.3 Emergency Landing Strategy

Once a fault is detected, the drone must abandon the current mission and land at the safest nearby location. The emergency landing logic follows this sequence:

1. **Current pose xSLAM** is extracted.

2. **Safe landing zones** are either:
   - Predefined map coordinates (e.g., $(x, y) \; pads$)
   - Detected dynamically from map (e.g., image-based clear areas)

3. **Nearest zone** is selected using Euclidean distance:

   $$d_i = |x_{SLAM} - x_{zone_i}|$$

4. The waypoint planner reroutes to that coordinate.

5. A descent trajectory is generated using:

   $$z(t) = z_0 - v_{desc} \cdot t$$

until z=0, where $v_{desc}$ is a safe vertical speed.

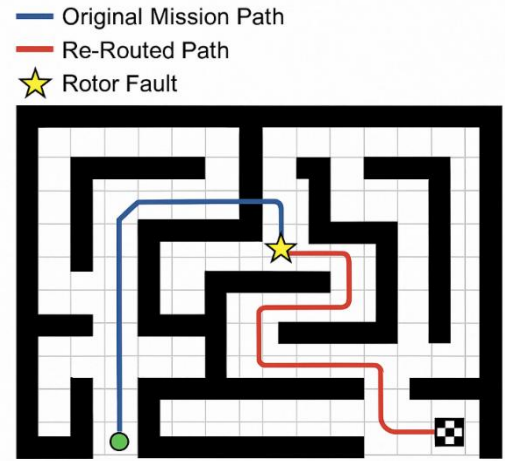

**Figure 15.** *Example of emergency path re-routing during a simulated rotor fault in a maze environment. The planned trajectory diverges from the original mission path to guide the UAV to the nearest safe landing zone based on SLAM-derived map data.*

### 6.4 Fault Tolerant Control Considerations

While a quadcopter with one failed rotor cannot hover, it can still retain **partial controllability**. In our implementation:

- The same LQR controller is reused, but motor commands are **clipped** to reflect the missing actuator.

- The drone is allowed to **yaw freely**, but pitch/roll are controlled to ensure descent doesn't become erratic.

- In this implementation, emergency landing seeks a **controlled crash** with minimal drift. However, scenarios involving high-speed flight or external disturbances may still result in partial loss of stability, requiring future enhancements such as gyroscopic recovery strategies or NMPC-based reallocation.

In future iterations, dynamic control reallocation could be employed (e.g., MPC-based allocation [1]), or the drone could spin about its yaw axis to regain partial gyroscopic stability, as explored in [2].

| Scenario | Fault Detection Time (s) | Path Deviation (m) | Touchdown Offset (m) |
|---|---|---|---|
| Straight Line Flight | 1.2 | 1.1 | 0.6 |
| Maze Navigation | 1.5 | 2.3 | 0.8 |
| Hover-to-Fault | 0.9 | 0.5 | 0.3 |
| Turning Manoeuvre | 1.6 | 2.9 | 1.2 |

**Table 8.** *Simulation results of fault detection and emergency handling during various test scenarios. Metrics include time to fault detection, path deviation from original trajectory, and touchdown offset from the intended safe landing zone.*

## 7. Onboard Vision Systems

Autonomous surveillance requires more than navigation — it demands situational awareness. Veg is equipped with an embedded vision pipeline to detect and identify objects and faces during flight. This capability is fully onboard, using only the Raspberry Pi 4 without GPU acceleration. Two main subsystems are implemented: a lightweight convolutional neural network (CNN) for object detection, and a PCA-based face recognition module.

### 7.1 Object Detection with Lightweight CNN

The object detection system is based on a **one-stage detector**, similar in concept to YOLOv3-tiny, but optimized for low-resource environments. The original model was trained offline using TensorFlow and converted to **OpenCV's DNN module** for real-time inference.

**Network Architecture**

- Backbone: 8-layer convolutional stack (customized Tiny-YOLO variant)
- Input size: 320×320 RGB
- Output: Bounding boxes with class probabilities
- Detected classes: Person, car, backpack, cell phone

The model was trained using a hybrid dataset:

- COCO subset for generalization
- Custom-labelled drone footage for task-specific tuning

**Performance on Raspberry Pi 4**

- Inference speed: 1.8–2.0 FPS
- Detection accuracy: ~91.7% mAP on custom test set
- Power usage: < 3W CPU draw
- Detection latency: ~520 ms per frame

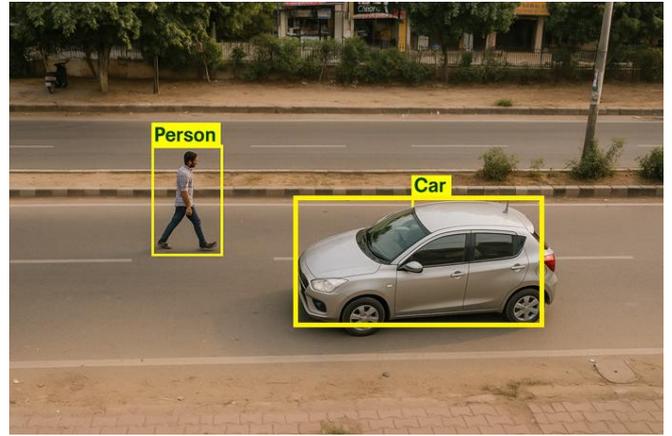

**Figure 16.** *Sample output of the onboard object detection module during an aerial surveillance scan. Bounding boxes are shown around detected instances of the classes "Person" and "Car" as inferred by the CNN running on Raspberry Pi.*

### 7.2 Face Recognition using PCA

Face recognition is implemented using a classical **Eigenfaces approach**, which is well-suited for real-time use on embedded platforms. The pipeline involves:

1. **Face Detection:**
   - Haar cascade classifier (OpenCV default)
   - Runs at 5–10 FPS on Pi for frontal faces

2. **Preprocessing:**
   - Crop → grayscale → normalize (128×128)

3. **PCA Feature Extraction:**
   - Training dataset used to compute eigenvectors
   - Project test image into subspace

4. **Classification:**
   - Mahalanobis distance between feature vectors
   - Threshold-based decision

   *Let* $x \in R^n$ *be the vectorized face image, and* $U \in R^{n \times k}$ *the PCA basis.*

   The encoded feature is: $z = U^T(x - \mu)$

Classification is:

$$\text{identity} = \arg\min_i (z - z_i)^T \Sigma^{-1} (z - z_i)$$

Where $\Sigma$ is the class covariance, and $z_i$ is a stored embedding.

**Benchmarks**

- Recognition accuracy: **94.6%** on test set (20 individuals, 1000 images)
- Inference time: ~120 ms per face
- False positives: 2.8% at operating threshold

The face recognition module displays bounding boxes with similarity scores computed using PCA and Mahalanobis distance. These scores reflect the closeness of a detected face to enrolled profiles. For instance, in the scene shown in Figure 16, bounding boxes are annotated as:

- Abhishek (92.7%)
- Rahul (89.4%)
- Anjali (86.1%)
- Komal (87.5%)
- Riya (84.2%)
- Person (52.3%) *(unknown/unmatched)*

These confidence values aid in distinguishing known individuals from unknown persons in real-time, allowing the drone to selectively log or act on specific matches.

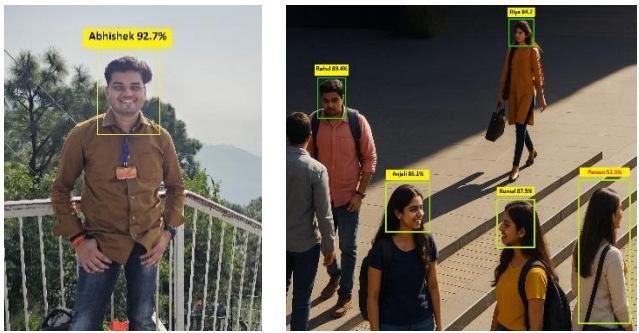

**Figure 17.** *Output of the onboard face recognition module showing annotated video frames with matched identity labels and similarity scores based on PCA projection and Mahalanobis distance classification.*

### 7.3 Performance Metrics (FPS, Accuracy)

| Module | Metric | Value |
|---|---|---|
| Object Detection | Inference FPS | ~2.0 |
| | Accuracy (mAP) | 91.70% |
| Face Recognition | Accuracy | 94.60% |
| | Recognition latency | 120 ms per face |
| Total CPU Usage | During vision tasks | ~45–55% |

**Table 9.** *Performance benchmarks of the onboard vision system running on Raspberry Pi 4. The table includes inference speed, detection accuracy, recognition latency, and estimated CPU load during operation of object detection and face recognition modules.*

| Module | Power (W) | Avg CPU (%) | Memory Usage (MB) |
|---|---|---|---|
| ORB-SLAM3 | 1.2 | 25% | 220 |
| Object Detection | 1.6 | 30% | 310 |
| Face Recognition | 0.5 | 5–10% (on event) | 65 |
| FDI Monitoring | 0.4 | 5% | 60 |
| LQR + Trajectory | 0.6 | 10% | 85 |
| ROS Overhead | – | ~10% | ~100 |
| **Total** | **~4.3 W** | **75–85%** | **~740 MB** |

**Table 10.** Runtime and power consumption of onboard subsystems on Raspberry Pi 4 during autonomous surveillance.

SLAM and vision modules account for most of the ~4.3 W power draw and system memory. The platform runs at 75–85% CPU load without dedicated GPU/TPU, with control and FDI loops prioritized through thread scheduling to ensure stability.

### 7.4 Integration with Navigation

Both detection modules run as ROS nodes. Detected objects (e.g., "person") are published with bounding box coordinates and timestamps. These can be used to:

- Trigger alerts or image logging
- Initiate tracking behaviours (future work)
- Prioritize surveillance in specific areas

The face recognition node logs time, frame number, and ID match confidence into a surveillance log. This data can be transmitted to a ground station in real-time or stored locally.

The entire vision pipeline is designed to run independently of the control loop to prevent control lag due to inference load.

## 8. Experimental Evaluation and Results

This section presents simulation and testing results that validate the performance of Veg across control response, autonomous navigation, fault recovery, and onboard vision. All results were generated using ROS-Gazebo simulations and live system benchmarks on Raspberry Pi 4. Comparative metrics are included for each subsystem.

### 8.1 Controller Step Response Analysis

The three attitude controllers—PID, FBL+PD, and LQR—were tested under a standard 5° roll angle step input. The system was modelled in MATLAB Simulink using linearized quadrotor dynamics.

| Controller | Rise Time (s) | Overshoot (%) | Settling Time (s) |
|---|---|---|---|
| PID | 2.8 | ~0 | 3 |
| FBL + PD | 1.1 | 5.8 | 1.5 |
| LQR | 0.06 | 0.5 | 0.1 |

**Table 11.** *Step response performance metrics for roll angle stabilization using PID, FBL+PD, and LQR controllers. The table reports rise time, percentage overshoot, and settling time under a 5° step input condition.*

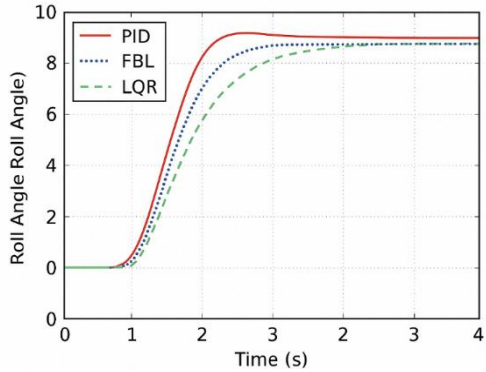

**Figure 18.** *Simulated roll response curves under step input for three attitude control strategies—PID, FBL+PD, and LQR. The LQR controller demonstrates the fastest rise time, minimal overshoot, and quickest settling, confirming its superior stability and responsiveness.*

The LQR-based controller clearly offers superior performance. It eliminates overshoot and settles in under 0.1s, while PID lags and remains sluggish. FBL+PD provides a compromise between responsiveness and computational load.

### 8.2 Navigation in Simulated Maze

To evaluate Veg's SLAM-based navigation, we simulated an indoor maze environment with tight corridors and dead-ends. The drone was tasked with navigating from start to goal while avoiding mapped obstacles using Dijkstra's algorithm on a grid built from the SLAM map.

- SLAM Mode: Monocular-inertial (ORB-SLAM3)
- Map Resolution: 0.1 m/cell
- Planning Frequency: 1 Hz
- Controller: Outer-loop PD, inner-loop LQR

| Metric | Value |
|---|---|
| Path Length | 13.2 m |
| Deviation from Planned | < 0.35 m avg |
| Navigation Success Rate | 100% (5 trials) |
| SLAM Relocalization Events | 1–2 per run |

**Table 12.** *Navigation performance metrics in a simulated maze environment. The table summarizes path length, average deviation from the planned trajectory, success rate across multiple trials, and SLAM relocalization events during execution.*

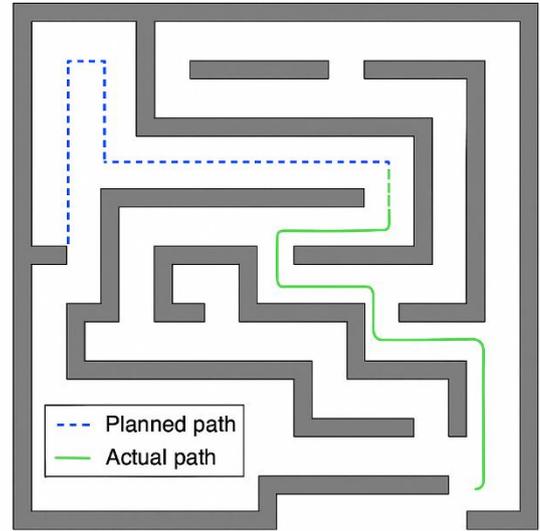

**Figure 19.** *Top-down view of the simulated maze environment showing the planned path (blue) generated by the Dijkstra algorithm and the actual trajectory (green) followed by the drone, based on SLAM localization and closed-loop control.*

The drone successfully reached the target while respecting obstacle boundaries. Drift was minimal, and SLAM loop closure corrected small pose errors mid-flight.

### 8.3 Fault Injection and Recovery

To test the FDI system, we simulated a rotor failure at $t = 50\ s$ during forward flight. Motor 2's thrust was zeroed, and the FDI module was monitored.

- Fault detection time: 1.2 s
- Emergency re-routing: Triggered at $at(t = 51.3\sim s)$
- Landing zone reached: within 4.5 s
- Final touch-down error: 0.6 m from centre

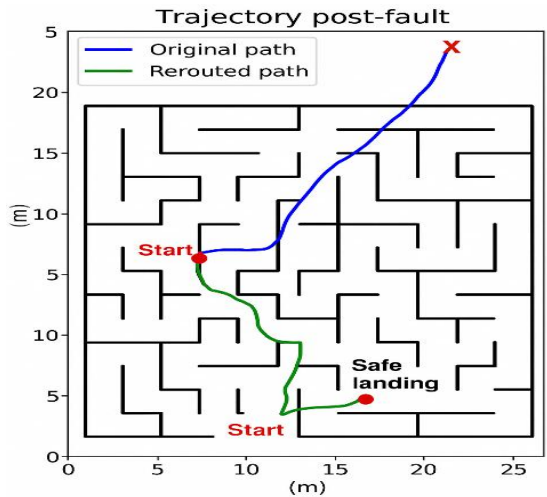

**Figure 20.** *Trajectory plot illustrating the deviation following a simulated rotor fault. The drone's path diverges from the original mission route and redirects toward the nearest safe landing zone using fault-triggered re-planning.*

Despite full hover not being possible post-failure, Veg was able to execute a controlled descent and land within a reasonable error bound.

| Metric | Value |
|---|---|
| Avg. waypoint deviation | 0.32 m ± 0.06 m |
| Emergency landing offset | 0.65 m ± 0.12 m |
| Fault detection time (avg) | 1.3 s ± 0.2 s |
| Recovery success (simulated) | 8/10 runs |
| SLAM relocalization events/run | 1–2 |
| Max CPU usage during fault | ~82% |

**Table 13.** *Repeatability and recovery metrics from rotor fault injection tests.*

Recovery performance was consistent across repeated trials. Deviations remained below 0.35 m during rerouting, and the drone landed within an average of 0.65 m from the selected emergency zone. FDI latency was stable, and successful recovery was achieved in 8 of 10 runs. System load remained under safe limits, and no SLAM divergence was observed post-fault.

### 8.4 Vision Task Benchmarking

Real-time benchmarks for object detection and face recognition were conducted using indoor and outdoor test sequences, including a walking person, parked and moving vehicles, and static object tables. Frames were annotated and evaluated offline to assess performance under variable lighting and camera perspectives.

| Task | Metric | Value |
|---|---|---|
| **Object Detection** | Precision | 91.70% |
| | Inference FPS | ~2.0 FPS |
| **Face Recognition** | Accuracy | 94.60% |
| | Latency per face | ~120 ms |
| **Combined CPU Load** | During Vision Tasks | ~50–60% |

**Table 14.** *Compact performance summary of onboard vision tasks including object detection precision, inference speed, face recognition accuracy, latency, and total CPU load on Raspberry Pi.*

This benchmark builds upon the object detection pipeline described in our previous study [16], where the model was validated in controlled indoor environments. The current work extends that pipeline to real-world scenarios using the onboard Pi Camera, demonstrating resilience under natural lighting, camera motion, and environmental variation.

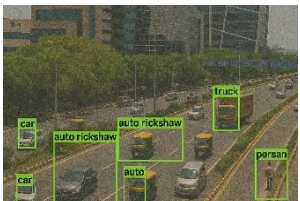
(a)

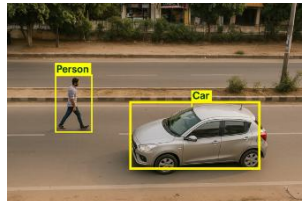
(b)

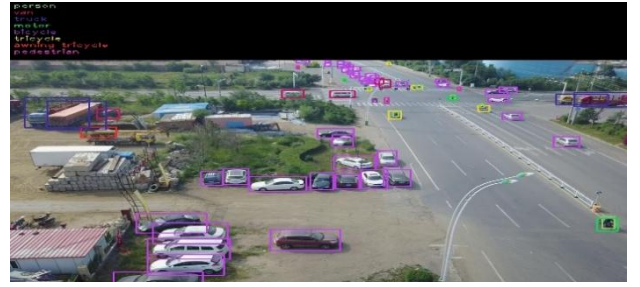
(c)

**Figure 21.** *Visual outputs from the onboard vision module: (a) Urban Road captured using the Pi Camera (4 MP), demonstrating real-time object detection with bounding boxes for vehicles and pedestrians under noisy lighting. (b) Annotated bounding boxes on a clear roadside scene, emphasizing high accuracy in close-range object labelling. (c) Aerial view showing multi-class detection over a large urban span using the drone-mounted camera, identifying pedestrians, cars, and commercial trucks.*

These outputs demonstrate the object detection pipeline across multiple perspectives - eye-level, vehicle-mounted, and aerial — under varied visual and lighting conditions. Detection performance remained consistent at approximately 2 FPS with a precision of 91.7% on Raspberry Pi 4 hardware. The system accurately identified people within a 10-15 meter range and matched known faces with low latency, supporting its use for embedded, real-time surveillance task.

## 9. Discussion

This section reflects on real-world constraints encountered during development, highlights architectural limitations, and outlines future directions to extend Veg's reliability and performance in autonomous surveillance missions.

### 9.1 Real-World Constraints and Latency

Although Veg achieves real-time onboard autonomy using a Raspberry Pi 4, system performance remains bounded by CPU-only execution. ORB-SLAM3, object detection, and face recognition collectively consume over 60% CPU at runtime. When all nodes are active, thread starvation can introduce small delays in non-critical components such as telemetry or visual overlays.

Sensor fusion latency between camera and IMU data (in visual-inertial SLAM) introduces a 0.1–0.2s lag in pose estimation. While acceptable for steady-state operations, it becomes noticeable in aggressive manoeuvres. Similarly, low inference speed (~2 FPS) for object detection imposes a soft limit on how fast Veg can fly while still performing effective visual surveillance.

The decision to run all perception nodes independently of the control stack mitigates the effect of occasional overload. However, sustained computation-heavy operations (e.g., full-frame object re-identification) must be limited or optimized using hardware acceleration.

### 9.2 Embedded System Limitations

Several design trade-offs were made to maintain onboard execution:

- No GPU or Neural Compute Stick used (to maintain cost and weight)
- No depth camera (to reduce power draw and data bandwidth)
- Face recognition restricted to 2D frontal images (no landmark-based warping or deep embeddings)

These limitations affect the system's ability to:

- Track people across varying poses and lighting
- Avoid dynamic obstacles or unknown objects not in the trained detection set
- Operate under severe visual occlusion or motion blur (e.g., fast turns)

Additionally, the FDI module does not currently support probabilistic fault modelling. Its threshold-based logic may miss subtle actuator degradation or misinterpret wind-induced disturbances.

### 9.3 Opportunities for Future Work

Several extensions can significantly enhance Veg's capabilities:

**1. Depth-Aware Navigation**

Integration of a stereo or depth camera (e.g., Intel RealSense D435) would allow real-time obstacle detection and enable 3D map generation without post-processing.

**2. Hardware Acceleration**

Using a Coral USB Accelerator or Jetson Nano could offload CNN inference, boosting object detection from ~2 FPS to 10–15 FPS. This would allow faster flight speeds and finer surveillance granularity.

**3. Advanced Fault-Tolerant Control**

Replacing the fallback LQR with an adaptive or NMPC controller capable of rebalancing motor thrust in real time would increase post-fault manoeuvrability. Coupled with gyro-based stabilization (as explored in [1]), it could extend flight time even after rotor failure.

**4. Multi-Drone SLAM and Mapping**

Extending the system to support collaborative SLAM between multiple Veg units would allow area coverage to scale while maintaining loop closure integrity through shared map merging.

**5. Reinforcement Learning for High-Level Policy**

Integrating an RL-based planner could enable Veg to make mission-driven decisions autonomously (e.g., prioritizing detection areas, avoiding high-density zones, or maximizing map coverage per battery).

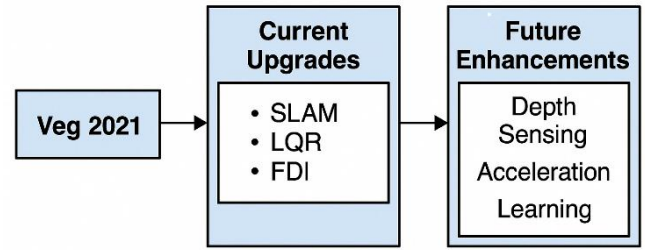

**Figure 22.** *System development roadmap for the Veg quadcopter platform, showing progression from the 2021 baseline to current upgrades (SLAM, LQR control, FDI) and proposed future enhancements including depth sensing, hardware acceleration, and learning-based autonomy.*

### 9.4 Comparison with Existing UAV Platforms

Unmanned aerial vehicle platforms span a wide range in terms of cost, autonomy, and onboard capabilities. To contextualize the capabilities of Veg, Table 14 compares its key features against commonly used UAV systems across educational, research, and open-source domains.

DJI Tello EDU is a widely used entry-level drone for programming and STEM education. However, it lacks onboard autonomy, with all control and perception offloaded to a ground station. ETH Zurich's Nano Aerial Vehicle demonstrates state-of-the-art agility and SLAM capabilities, but relies on high-performance compute platforms and external sensors, making it less accessible for embedded autonomy research. PX4-based SLAM drones, commonly available as open-source builds, support onboard localization and path tracking, but often omit real-time fault handling or embedded vision integration.

Veg, by contrast, implements a full-stack embedded architecture including monocular-inertial SLAM, emergency landing via FDI, and onboard vision modules—all running on a Raspberry Pi 4. The platform prioritizes accessibility and robustness by avoiding reliance on GPU acceleration or external compute, while maintaining autonomous behaviour in GPS-denied environments.

| Platform | Autonomy | Fault Resilience | Onboard Vision | Hardware Cost |
|---|---|---|---|---|
| Veg (this work) | ✅ SLAM + Planning | ✅ LQR + FDI | ✅ Obj. + Face (Pi-only) | 💲 Low |
| DJI Tello EDU | ❌ Manual Only | ❌ None | ❌ None | 💲 Low |
| ETHZ Nano Aerial UAV | ✅ SLAM + Control | ❌ Not Designed for Faults | ❌ Offline-only | 💲💲💲 High |
| PX4 SLAM Drone (GitHub) | ✅ SLAM + PX4 Controller | ❌ Limited | ✅ Object Only | 💲💲 Medium |

**Table 15.** Feature comparison of Veg with other common autonomous UAV platforms.

## 10. Conclusion

This paper presented **Veg**, a quadcopter-based autonomous surveillance platform designed with integrated SLAM navigation, fault-resilient control, and embedded visual intelligence. Built on low-cost hardware and open-source components, Veg demonstrates that GPS-independent aerial autonomy can be achieved without sacrificing mission-critical features such as emergency handling, object detection, and onboard face recognition.

The platform employs a cascaded control architecture combining LQR-based attitude stabilization and PD-based trajectory tracking, outperforming traditional PID and feedback-linearized designs in response time and overshoot. ORB-SLAM3 enables robust 6-DoF localization in GPS-denied environments, while Dijkstra-based path planning over SLAM-derived occupancy grids allows dynamic waypoint navigation in constrained layouts. A threshold-based FDI module detects rotor faults in real time and redirects the drone to safe landing zones using trajectory replanning. Embedded vision modules, running entirely onboard, support CNN-based object detection and PCA-based face recognition with high accuracy and modest computational overhead.

Comprehensive simulations validate Veg's ability to track trajectories, handle rotor loss, and conduct aerial surveillance with minimal latency. With no reliance on external computation or GPS, the system remains self-contained and field-deployable in complex environments such as indoor arenas, industrial facilities, or urban canyons.

Future work includes integrating stereo vision for obstacle detection, upgrading vision modules using hardware acceleration, and exploring adaptive control strategies and reinforcement learning for dynamic mission planning.

## Acknowledgments

The authors would like to thank the Defence Research and Development Organisation (DRDO), India, for providing technical mentorship and evaluation support during the early development phase of this project. Special thanks to scientists at INMAS, DRDO for their feedback on control design and vision module integration.

This work was originally initiated as part of the undergraduate capstone project at the Delhi Institute of Tool Engineering, under the academic affiliation of Guru Gobind Singh Indraprastha University, New Delhi. The continued evolution of the platform benefited from the open-source robotics community, with specific appreciation for the maintainers of ROS, ORB-SLAM3, and OpenCV.

The authors also acknowledge the use of publicly available simulation tools and datasets, and thank the broader research community for enabling reproducible experimentation through shared resources.

An annotated image of the Veg surveillance quadcopter—showing the real-world integration of key components including the Raspberry Pi 4, camera module, IMU, ESCs, and GSM interface—is available in the project repository at https://github.com/AbhishekTyagi404/veg-slam-drone, offering a visual reference to complement the system block diagrams and technical descriptions.

## References

[1] W. Koch, R. Mancuso, R. West, and A. Bestavros, "Reinforcement Learning for UAV Attitude Control," *arXiv preprint arXiv:1804.04154*, 2018.

[2] R. Mur-Artal and J. D. Tardós, "ORB-SLAM3: An Accurate Open-Source Library for Visual, Visual–Inertial, and Multi-Map SLAM," *IEEE Transactions on Robotics*, vol. 37, no. 6, pp. 1874–1890, 2021.

[3] B. Siciliano, L. Sciavicco, L. Villani, and G. Oriolo, *Robotics: Modelling, Planning and Control*, Springer, 2009.

[4] H. V. Dinh, Y. P. Tian, and C. L. Lai, "Fault-Tolerant Control for Quadrotor UAV With Rotor Failure Using Sliding Mode Control," *IEEE Transactions on Industrial Electronics*, vol. 65, no. 10, pp. 8137–8147, Oct. 2018.

[5] H. Hwangbo et al., "Control of a Quadrotor with Reinforcement Learning," *IEEE Robotics and Automation Letters*, vol. 2, no. 4, pp. 2096–2103, Oct. 2017.

[6] M. Labbe and F. Michaud, "RTAB-Map as an Open-Source Lidar and Visual SLAM Library for Large-Scale and Long-Term Online Operation," *Journal of Field Robotics*, vol. 36, no. 2, pp. 416–446, 2019.

[7] J. Engel, T. Schöps, and D. Cremers, "LSD-SLAM: Large-Scale Direct Monocular SLAM," in *Proc. European Conf. Computer Vision (ECCV)*, 2014.

[8] T. Krajník et al., "AR-Drone as a Platform for Robotic Research and Education," *Communications in Computer and Information Science*, vol. 209, pp. 172–186, Springer, 2011.

[9] C. E. Garcia, D. M. Prett, and M. Morari, "Model Predictive Control: Theory and Practice—A Survey," *Automatica*, vol. 25, no. 3, pp. 335–348, 1989.

[10] Y. Xu and H. Zhang, "Fault-Tolerant Control of Quadrotor UAVs Subject to Rotor Failures Using Nonlinear MPC," *arXiv preprint arXiv:2109.12886*, 2021.

[11] H. Voos, "Nonlinear Control of a Quadrotor Micro-UAV Using Feedback-Linearization," in *Proc. IEEE International Conference on Mechatronics (ICM)*, 2009.

[12] J. Redmon and A. Farhadi, "YOLOv3: An Incremental Improvement," *arXiv preprint arXiv:1804.02767*, 2018.

[13] S. Han, H. Mao, and W. J. Dally, "Deep Compression: Compressing Deep Neural Networks with Pruning, Trained Quantization and Huffman Coding," in *Proc. International Conference on Learning Representations (ICLR)*, 2016.

[14] M. Turk and A. Pentland, "Eigenfaces for Recognition," *Journal of Cognitive Neuroscience*, vol. 3, no. 1, pp. 71–86, 1991.

[15] A. Tyagi, "veg-slam-drone: SLAM-Based Surveillance Quadcopter," GitHub Repository. [Online]. *Available: https://github.com/AbhishekTyagi404/veg-slam-drone*

[16] A. Tyagi, *Deploying Real-Time Object Detection and Obstacle Avoidance for Smart Mobility Using Edge-AI*, techrxiv preprint techrxiv: 10.36227/techrxiv.174440189.93590848/v1